\documentclass[manuscript,screen]{acmart}

\renewcommand\footnotetextcopyrightpermission[1]{} 
\usepackage{subfig}
\usepackage{algorithm}
\usepackage{algorithmic}
\usepackage{multirow}
\newcommand{\sign}{\text{sign}}
\AtBeginDocument{%
  \providecommand\BibTeX{{%
    \normalfont B\kern-0.5em{\scshape i\kern-0.25em b}\kern-0.8em\TeX}}}

\begin{document}

\title{Block Walsh-Hadamard Transform Based Binary Layers in Deep Neural Networks}

\author{Hongyi Pan}
\email{hpan21@uic.edu}
\orcid{0000-0001-9421-3936}
\author{Diaa Badawi}
\email{dbadaw2@uic.edu}
\author{Ahmet Enis Cetin}
\email{aecyy@uic.edu}
\affiliation{%
  \institution{Department of Electrical and Computer Engineering, University of Illinois Chicago}
  \streetaddress{1200 W Harrison St}
  \city{Chicago}
  \state{Illinois}
  \country{USA}
  \postcode{60607}
}


\begin{abstract}
Convolution has been the core operation of modern deep neural networks. It is well-known that convolutions can be implemented in the Fourier Transform domain.  In this paper, we propose to use binary block Walsh-Hadamard transform (WHT)  instead of the Fourier transform. We use WHT-based binary layers to replace some of the regular convolution layers in deep neural networks. We utilize both one-dimensional (1-D) and two-dimensional (2-D) binary WHTs in this paper.
  In both 1-D and 2-D layers, we compute the binary WHT of the input feature map and denoise the WHT domain coefficients using a nonlinearity which is obtained by combining soft-thresholding with the tanh function. After denoising, we compute the inverse WHT. We use 1D-WHT to replace the $1\times 1$ convolutional layers, and 2D-WHT layers can replace the 3$\times$3 convolution layers and Squeeze-and-Excite layers. 2D-WHT layers with trainable weights can be also inserted before the Global Average Pooling (GAP) layers to assist the dense layers. In this way, we can reduce the number of trainable parameters significantly with a slight decrease in trainable parameters.  In this paper, we implement the WHT layers into MobileNet-V2, MobileNet-V3-Large, and ResNet to reduce the number of parameters significantly with negligible accuracy loss. Moreover, according to our speed test, the 2D-FWHT layer runs about 24 times as fast as the regular $3\times 3$ convolution with 19.51\% less RAM usage in an NVIDIA Jetson Nano experiment.

\end{abstract}
\thanks{This work is funded by an award from the University of Illinois Chicago Discovery Partners Institute Seed Funding Program and NSF grants 1739396 and 1934915.}



\maketitle

\section{Introduction}
Recently, deep convolution neural networks (CNNs) have enjoyed a great success in many important applications such as image classification~\cite{krizhevsky2012imagenet,simonyan2014very, szegedy2015going, wang2017residual, he2016deep, he2016identity, badawi2020computationally, agarwal2021coronet, partaourides2020self, stamoulis2018designing}, object detection~\cite{redmon2016you, aslan2020deep, menchetti2019pain, aslan2019early} and semantic segmentation~\cite{yu2018bisenet, huang2019ccnet, long2015fully, poudel2019fast, jin2019fast}. While more and more parameters are needed in deep neural networks, deploying them on real-time resource-constrained environments such as embedded devices becomes a difficult task due to insufficient memory and limited computational capacity. To overcome the aforementioned limitations, smaller and computationally efficient neural networks are necessary for many practical applications.

Although $1\times1$ convolutions reduce the computational load, they are still computationally expensive and time-consuming in regular deep neural networks. In our previous paper \cite{pan2021fast}, we proposed a binary layer based on Fast Walsh-Hadamard transform (FWHT) to replace the $1\times1$ convolution layer. We revised MobileNet-V2~\cite{sandler2018mobilenetv2} with the FWHT layer, and the new network is remarkably more slimmed and computationally efficient compared to the original structure according to our experiments. However, we can take advantage of the fast $O(m\log_2m)$ Walsh-Hadamard Transform (WHT) algorithm only when $m$ is an integer power of 2. As a result, we pad zeros to the end of the input vector in the FWHT layer to make the size of the vector an integer power of 2. We notice that padding zeros to increase the size of the Walsh-Hadamard Transform (WHT) brings some redundant parameters. In this paper, we introduce both one-dimensional (1D) and 2-Dimensional (2D) Walsh-Hadamard Transform layers. The new version of the 1D WHT computes the WHT in small blocks and avoids the zero-padding operations. As a result it needs fewer parameters compared to our earlier paper \cite{pan2021fast}. Our 2D FWHT layer can replace the $3\times3$ convolution layers and Squeeze-and-Excite layers. It can also be employed to assist the dense layers of a given network. 
Our contribution can be summarized as follows:

\begin{itemize}
\item We have a new 1D-FWHT layer. It improves the results of our previous work~\cite{pan2021fast} by using Blocks of Walsh-Hadamard Transform (BWHT) instead of a single WHT computation to replace an entire $1\times1$ convolution layer. Compared to the old 1D-FWHT layer, the new 1D-BWHT layer retains the accuracy loss with a little more parameter reduction because it eliminates the zero-padding operation required to make the transform size a power of 2. For example, our version of the MobileNet-V2 network with the BWHT layers reaches a 0.08\% higher accuracy with 11,919 fewer parameters compared to the MobileNet-V2 network with the FWHT layer on the CIFAR-10 dataset.

\item We introduce the two-dimensional (2D) Walsh-Hadamard Transform layer in this paper. It can be used to replace the “Squeeze-and-Excite” layers and the regular 2D convolutional layers. For example, we reduce the number of trainable parameters by 48.62\% with only a 0.76\% accuracy loss in our MobileNet-V3-Large structure with 2D-WHT layers on the CIFAR-100 dataset. Our version of RESNET-34 with 2D-WHT layers has 53.57\% fewer parameters than the regular RESNET-34 with only a 0.72\% accuracy loss on the tiny ImageNet dataset.

\item In addition, we introduce a weighted 2D-FWHT layer that can be easily inserted before the global average pooling (GAP) layer (or the flatten layer) to assist the dense layers. This novel layer improves the accuracy of the network with a slight (almost negligible) increase in parameters due to the additional weights. For example, our additional weighted 2D-FWHT layer improves the accuracy of the ResNet-20 network by 0.5\% with only 256 additional parameters in the CIFAR-10 dataset.
    
This 2D-FWHT layer improves the accuracy of the MobileNet-V3-Large network by 0.15\% in the CIFAR-100 dataset. In this case, only 704 more parameters are required. 

Our novel binary layers do not increase the processing time even in conventional processors. For example, our 2D-FWHT layer runs about 24 times as fast as the regular $3\times3$ convolution layer on NVIDIA Jetson Nano.

\end{itemize}

\section{Related Work}
Efficient neural network models include compressing a large neural network using quantization~\cite{wu2016quantized,muneeb2020robust}, hashing~\cite{chen2015compressing}, pruning~\cite{pan2020computationally}, vector quantization~\cite{yu2018gradiveq} and Huffman encoding~\cite{han2019deep}. Another approach is the  SqueezeNet~\cite{iandola2016squeezenet}, which is designed as a small network with $1\times 1$ convolutional filters. Yet another approach is to use binary weights in neural networks \cite{courbariaux2016binarized, bulat2018hierarchical, rastegari2016xnor, shen2021s2, liu2020reactnet, martinez2020training, bulat2020bats, hubara2016binarized, alizadeh2018empirical, bannink2020larq, juefei2017local, lin2017towards, wang2019learning, zhao2019building}. Since the weights of the neurons are binary, they can can be used to slim and accelerate networks in specialized hardware including compute-in-memory systems~\cite{nasrin2021mf}. 

In 2020, Maneesh \textit{et al.} proposed a trimmed version of MobileNet architecture called Reduced Mobilenet-V2~\cite{ayi2020rmnv2}. They replace bottleneck layers with heterogeneous kernel-based convolutions (HetConv) blocks. HetConv was first proposed by Pravendra \textit{et al.} in 2019 \cite{singh2019hetconv}. Although HetConv reduced the parameters effectively, it is still based on the convolution, and we experimentally showed that our approach can outperform the network described in \cite{ayi2020rmnv2} .

In 2021, James \textit{et al.} proposed a novel layer based on the Fast Fourier Transform (FFT) called FNet for natural language processing applications. They insert the FFT in hidden layers and they can train the weights in the frequency domain (Fourier domain). They showed that their FNet has a faster speed with the guaranteed accuracy on BERT counterparts on the GLUE benchmark. However, the main weakness  of the FNet is that their proposed method is only based on the real part of the FFT. After applying the FFT, they only keep the real part to avoid complex arithmetic. The imaginary part of the FFT is simply ignored. Therefore, the information in the imaginary part is lost. Since the WHT is a real transform we do not suffer from information loss as a result of WHT computation. 

Other methods using the Hadamard transform include~\cite{deveci2018energy, zhao2019building} but they did not perform any "convolutional" and non-linear filtering in the Hadamard transform domain. The "convolutional" filtering in the transform domain is possible by introducing multiplicative weights and non-linear filtering is possible with the use of two-sided
 smooth-thresholding which "denoises" small valued transform domain coefficients \cite{pan2021fast}. The use of one-sided RELU will lead to a significant information loss because transform domain coefficients can take both positive and negative values even if the input to the transform consists of all positive numbers.  Our novel smooth thresholding nonlinearity is a two-sided version of the RELU and it can retain both positive and negative large amplitude coefficients while eliminating small amplitude ones. After performing the filtering operations in the transform domain we compute the inverse transform and continue processing the data in the feature map domain, which is a unique feature of our paper.

\section{Review of the Fast Walsh-Hadamard (FWH) Transform}
    The Walsh-Hadamard Transform (WHT) is an example of a generalized class of the Fourier transforms. The transform matrix consists of +1 and -1 only. The
    convolution in the time (feature) domain leads to multiplication in the frequency domain in Fourier Transform. The Hadamard transform can be considered as a simplified version of the Fourier and wavelet transforms \cite{cetin1993block}. Therefore, we can approximately implement  $1\times1$ and $3\times3$ convolutions in the WHT domain. We can not only train multiplicative weights  but also train the threshold values of the soft and smooth-thresholds using the backpropagation algorithm \cite{pan2021fast}.  We experimentally observed that this approach approximates $1\times1$ and $3\times3$ convolution operations very effectively while significantly reducing the number of network parameters. In addition, we used trainable weights in 2D Hadamard transform layers (Section \ref{sec: 2D Weighted Walsh-Hadamard Transform layer}) and they positively contributed to the accuracy as shown in Table~\ref{tab: MobileNetV3_CIFAR100} in Section \ref{sec: 2D-FWHT in MobileNet-V3} and Tables \ref{tab: ResNet20_CIFAR10}m \ref{tab: ResNet34_TinyImageNet} in Section \ref{sec: 2D-FWHT in ResNet}.
    
    Let $\mathbf{X}, \mathbf{Y} \in \mathbb{R}^m$ be the vectors in the "time" and transform domains, respectively, where $m=2^k, k \in \mathbb{N}$. The WHT vector $\mathbf{Y}$  is obtained from $\mathbf{X}$ via a matrix multiplication as follows:
    \begin{equation}
        \mathbf{Y} = \mathbf{W}_k\mathbf{X}
    \end{equation}
   where $\mathbf{W}_k$ is called the $2^k$-by-$2^k$ Walsh matrix, which can be generated using the following steps \cite{walsh1923closed}:
    \begin{enumerate}
        \item Construct the Hadamard matrix $\mathbf{H}_k$:
        \begin{equation}
			\mathbf{H}_k = 
			\begin{cases}
				1,& k = 0,\\
				\begin{bmatrix}
					\mathbf{H}_{k-1} & \mathbf{H}_{k-1} \\ \mathbf{H}_{k-1} & -\mathbf{H}_{k-1}
				\end{bmatrix},& k > 0,
			\end{cases}
		\end{equation}
		Alternatively, for $k>1$, $\mathbf{H}_k$ can also be computed using Kronecker product $\otimes$:
		\begin{equation}
			\mathbf{H}_k=\mathbf{H}_1 \otimes\mathbf{H}_{k-1}.
		\end{equation}
        \item Shuffle the rows of $\mathbf{H}_k$ to obtain $\mathbf{W}_k$ by applying the bit-reversal permutation and the Gray-code permutation on row index. 
    \end{enumerate}
    
  For $k=2$ and  $m=4$,
		the Hadamard matrix		$\mathbf{H}_2 $	and the corresponding Walsh matrix $	\mathbf{W}_2$ are given by
		\begin{equation}
			\mathbf{H}_2 =
			\begin{bmatrix}
				1&1&1&1\\1&-1&1&-1\\1&1&-1&-1\\1&-1&-1&1\\
			\end{bmatrix}, \ \ \ \ 
			\mathbf{W}_2 =
			\begin{bmatrix}
				1&1&1&1\\1&1&-1&-1\\1&-1&-1&1\\1&-1&1&-1\\
			\end{bmatrix}.
		\end{equation}
which can be implemented using a wavelet filterbank with binary filters $h[n] = \{ 1, 1 \}$ and $g[n] = \{ -1, 1 \}$ in two stages \cite{cetin1993block}. This process is the basis of the fast algorithm and the 	
$\mathbf{W}_2$ matrix can be expressed as follows
	\begin{equation}
			\mathbf{W}_2 =
			\begin{bmatrix}
				1&1&0&0\\1&-1&0&0\\0&0&1&-1\\0&0&1&1\\
			\end{bmatrix} \times 
			\begin{bmatrix}
				1&1&0&0\\0&0&1&1\\1&-1&0&0\\0&0&1&-1\\
			\end{bmatrix}
		\end{equation}
Similar to the Fast Fourier Transform (FFT) algorithm, the complexity of the Fast Walsh-Hadamard transform (FWHT) algorithm is also $O(m\log_2m)$ when $m$ is an integer power of 2 as it is completely based on the butterfly operations described in Eq. (1) in~\cite{fino1976unified}. Because the Walsh matrix only contains +1 and -1, the implementation of FWHT can be achieved using only addition and subtraction operations via butterflies. It was shown that the WH transform is the same as the block Haar wavelet transform in \cite{cetin1993block}.
	As for the inverse transform, we have
	\begin{equation}
	   \mathbf{X} = \frac{1}{m}\mathbf{W}_k\mathbf{Y},
	\end{equation}
    which implies that the inverse Walsh-Hadamard transform is itself with normalization by $m$.

\section{Methodology}
    In this section, we will first review our previous work on Fast Walsh-Hadamard Transform (FWHT) layer 
    \cite{pan2021fast}, then we describe the Block Walsh-Hadamard Transform (BWHT) layer. Next, we will introduce a novel 2D-Walsh-Hadamard Transform layer which can be easily inserted into any deep network before the Global Average Pooling (GAP) layer and/or the flatten layer. 
    We will also describe how it can replace the $3\times3$ convolution layers and the Squeeze-and-Excite layers to reduce the number of parameters significantly. Finally, we describe how we introduce multiplicative weights in the 2D WHT domain. Addition of weights in the 2D WHT transform domain causes a minor increase in the number of parameters but returns a higher accuracy than the regular 2D WHT layer.
    
    \subsection{Fast Walsh-Hadamard Transform Layer}
	\label{FWHT Layer}
	The $1\times1$ convolution layer provides amazing benefit in modern deep convolution layers~\cite{he2016deep, he2016identity, sandler2018mobilenetv2, xie2017aggregated, szegedy2017inception}. It is widely used to change the dimensions of channels. For example, in ResNet~\cite{he2016deep}, a $1\times1$ convolution layer is applied to the residual blocks when the input and output sizes are different. The $1\times1$ convolution layer named ``conv\_expand" is applied to increase the number of channels in each bottleneck layer of MobileNet-V2~\cite{sandler2018mobilenetv2}.
	Then, a depthwise convolution layer is employed as the main component for feature extraction. After this step, another $1\times1$ convolution layer named ``conv\_projection" reduces the number of channels. 
	
	However, there are as many parameters as the number of input channels in each $1\times1$ convolution layer. Computing these operations is very time-consuming during the inference. Since the main duty of the $1\times1$ convolution layer is just to adjust the number of channels we proposed a novel FWHT layer to replace the $1\times1$ convolution layer in \cite{pan2021fast}. The FWHT layer is summarized in Algorithms~\ref{al: FWHT layer expand} and~\ref{al: FWHT layer project}. In brief, an FWHT layer consists of an FWHT operation to change the input tensor to the WHT domain, a smooth-thresholding operation as the non-linearity function in the WHT domain, and another inverse FWHT to change the tensor back to the feature-map domain. Each FWHT is applied among the channel axis, which implies that completing FWHT on a tensor $\mathbf{X}\in\mathbb{R}^{n\times w\times h\times m}$ means performing $n\times w\times h$ $m$-length FWHTs in parallel.  
	
	We apply "denoising" in the WH transform domain to eliminate small-amplitude coefficients. This is a nonlinear filtering operation in the transform domain. Inspired by soft-thresholding \cite{agante1999ecg, donoho1995noising} which is defined as 
	\begin{equation}
		y = \text{S}_T (x) = \sign(x)(|x|-T)_+ = \begin{cases}
			x+T, & x < -T\\
			0, & |x| \le T\\
			x-T, & x > T
		\end{cases},
	\end{equation}
	we have proposed the variant called smooth-thresholding~\cite{pan2021fast}
	\begin{equation}
		y = \text{S}_T' (x) = \tanh(x)(|x|-T)_+,
	\end{equation}
	where $T$ is the thresholding parameter and is trainable in our FWHT layer. 
	
Due to its definition given in Eq.~(\ref{eq: dstdT}), the denoising parameter $T$ in soft-thresholding can only be updated by either $+1$ or $-1$. On the other hand, the derivative in Eq.~(\ref{eq: dttdT}) is the derivative in Eq.~(\ref{eq: dstdT}) multiplied by $\tanh(x)$. As a result, the convergence of the smooth-thresholding operator is smooth and steady in the back-propagation algorithm. 
	\begin{equation}
		\frac{\partial(\sign(x)(|x|-T)_+)}{\partial T} = \begin{cases}
			1, & x < -T\\
			0, & |x| \le T\\
			-1, & x > T
		\end{cases}
		\label{eq: dstdT}
	\end{equation}
	
	\begin{equation}
		\frac{\partial(\tanh(x)(|x|-T)_+)}{\partial T} = \begin{cases}
			-\tanh(x), & |x| > T\\
			0, & |x| \le T\\
		\end{cases}
		\label{eq: dttdT}
	\end{equation}
	
	We learn a different threshold $T$ value for each WHT domain coefficient. 
	
	We do not use the ReLU function in the transform domain because the WHT domain coefficients can take both positive and negative values, and large positive and negative transform domain coefficients are equally important. In our MobileNet-V2 experiments in \cite{pan2021fast} we have verified this observation, and both soft-thresholding and smooth-thresholding improve the recognition accuracy compared to the ReLU.
	
	
	Two types of the FWHT layer are summarized in Algorithms~\ref{al: FWHT layer expand} and~\ref{al: FWHT layer project}. Because the Direct Current (DC) channel $\mathbf{Y}[:, :, :, 0]$ usually contains essential information about the mean value of the input feature map, we do not apply any thresholding on it. If we perform smooth-thresholding on tensor $\mathbf{Y}$, each slice among the channel axis will share a common threshold value. Therefore, there are total $(2^d-1)$ trainable thresholding parameters in the $2^d$-channel FWHT layer.
	
	\begin{algorithm}[htbp]
		\caption{The FWHT layer for channel expansion \cite{pan2021fast}}
		\begin{algorithmic}[1]
			\renewcommand{\algorithmicrequire}{\textbf{Input:}}
			\renewcommand{\algorithmicensure}{\textbf{Output:}}
			\REQUIRE Input tensor $\mathbf{X}\in\mathbb{R}^{n\times w\times h\times c}$
			\ENSURE  Output tensor $\mathbf{Z}\in\mathbb{R}^{n\times w\times h\times tc}$
			\STATE Find minimum $d\in \mathbb{N}$, s.t. $2^d\ge tc$ 
			\STATE $\hat{\mathbf{X}} = \text{pad}(\mathbf{X}, 2^d-c) \in\mathbb{R}^{n\times w\times h\times 2^d}$
			\STATE $\mathbf{Y} = \text{FWHT}(\hat{\mathbf{X}}) \in\mathbb{R}^{n\times w\times h\times 2^d}$
			\STATE $\hat{\mathbf{Y}} = \text{concat}(\mathbf{Y}[:, :, :, 0], \text{ST}(\mathbf{Y}[:, :, :, 1:]))$
			\STATE $\hat{\mathbf{Z}} = \text{FWHT}(\hat{\mathbf{Y}}) \in\mathbb{R}^{n\times w\times h\times 2^d}$
			\STATE $\mathbf{Z} = \hat{\mathbf{Z}}[:, :, :tc]$
			\RETURN $\mathbf{Z}$.\\
			Comments: Function pad($\mathbf{A}, b$) pads $b$ zeros on the channel axis of tensor $\mathbf{A}$. FWHT($\cdot$) is the normalized fast Walsh-Hadamard transform on the last axis. Function concat($\cdot, \cdot$)  concatenates two tensors along the last axis. ST($\cdot$) performs smooth-thresholding. Index follows Python's rule.
		\end{algorithmic}
		\label{al: FWHT layer expand}
	\end{algorithm}
	
	\begin{algorithm}[htbp]
		\caption{The FWHT layer for channel projection \cite{pan2021fast}}
		\begin{algorithmic}[1]
			\renewcommand{\algorithmicrequire}{\textbf{Input:}}
			\renewcommand{\algorithmicensure}{\textbf{Output:}}
			\REQUIRE Input tensor $\mathbf{X}\in\mathbb{R}^{n\times w\times h\times tc}$
			\ENSURE  Output tensor $\mathbf{Z}\in\mathbb{R}^{n\times w\times h\times c}$
			\STATE Find minimum $p, q\in \mathbb{N}$, s.t. $2^p\ge tc, 2^q \ge c$ 
			\STATE $r = 2^{p-q}$
			\STATE $\hat{\mathbf{X}} = \text{pad}(\mathbf{X}, 2^p-tc) \in\mathbb{R}^{n\times w\times h\times 2^p}$
			\STATE $\mathbf{Y} = \text{FWHT}(\hat{\mathbf{X}}) \in\mathbb{R}^{n\times w\times h\times 2^p}$
			\STATE $\hat{\mathbf{Y}} = \text{concat}(\mathbf{Y}[:, :, :, 0]/r, \text{avgpool}(\text{ST}(\mathbf{Y}[:, :, :, 1:2^p-r+1]), r))\in\mathbb{R}^{n\times w\times h\times 2^q}$
			\STATE $\hat{\mathbf{Z}} = \text{FWHT}(\hat{\mathbf{Y}}) \in\mathbb{R}^{n\times w\times h\times 2^q}$
			\STATE $\mathbf{Z} = \hat{\mathbf{Z}}[:, :, :c]$
			\RETURN $\mathbf{Z}$.\\
			Comments: Function pad($\mathbf{A}, b$) pads $b$ zeros on the channel axis of tensor $\mathbf{A}$. FWHT($\cdot$) is the normalized fast Walsh-Hadamard transform on the last axis. Function concat($\cdot, \cdot$) concatenates two tensors along the last axis. Function avgpool($\mathbf{A}, b$) is the average pooling on $\mathbf{A}$ with pooling size and strides are $b$. ST($\cdot$) performs smooth-thresholding. Index follows Python's rule.
		\end{algorithmic}
		\label{al: FWHT layer project}
	\end{algorithm}
	
	To expand the number of channels, we first compute the $2^d$ point WHTs and perform smooth-thresholding in the transform domain. We pad $(2^d-c)$ zeros to the end of each input vector before the $2^d$-by-$2^d$ WHT to increase the dimension. After smooth-thresholding in the WH domain, we calculate the inverse WH transform. 

	To project channels by a factor of $r= \frac{2^p}{2^q}$, we first compute the $2^p$ point WHTs and perform smooth-thresholding in the transform domain as described in Algorithm \ref{al: FWHT layer project}. After this step, we compute the $2^q$ point WH transforms to reduce the dimension of the feature map. We divide the DC channel values by $r$ to keep the energy at the same level as other channels after pooling. In Step 5 of Algorithm \ref{al: FWHT layer project}, we average pool the transform domain coefficients to reduce the dimension of the WHT and discard the last $(r-1)$ transform domain coefficients of $\mathbf{Y}$ to make the dimension equal to $2^q$. The last $(r-1)$ coefficients are high-frequency coefficients, and usually, their amplitudes are negligible compared to other WHT coefficients.
	
\begin{figure}[htbp]
		\begin{center}
			\subfloat[FWHT for Channel Expansion]{\includegraphics[width=0.35\linewidth]{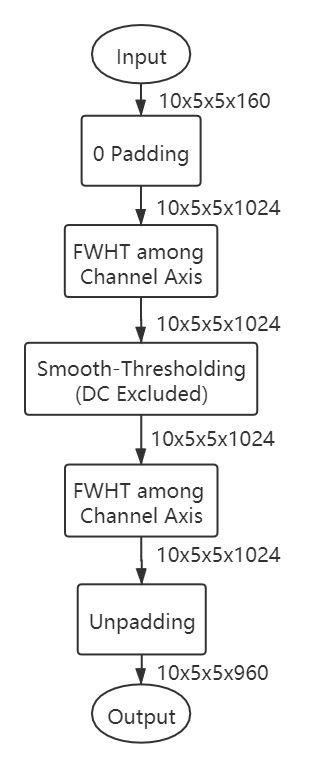}}
			\subfloat[FWHT for Channel Projection]{\includegraphics[width=0.35\linewidth]{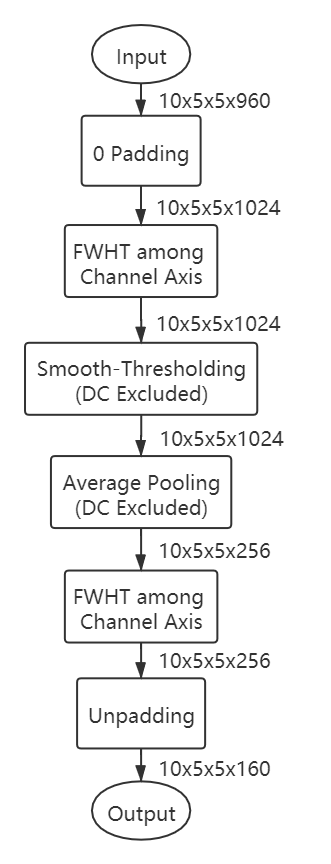}}
		\end{center}
		\caption{The FWHT layer (left) increases the number of channels by 6, and the FWHT layer (right) decreases the number of channels of a tensor by $\frac{1}{6}$, respectively. We pad zeros to the tensor to 1024 channels because 1024 is the minimum power of 2 greater than 960. There are 1023 trainable parameters in each FWHT layer.}
		\label{fig: fwht}
	\end{figure}

	Therefore, the dimension change operation from $m$ dimensions to $n$ dimensions can be summarized as follows:
	\begin{equation}
		\mathbf{Z} = \begin{cases}
			\frac{1}{2^{q}}\mathbf{U}\mathbf{W}_q\mathbf{S'_T}\mathbf{W}_q\mathbf{P}\mathbf{X},  \ \ \ m \le n\\
			\frac{1}{\sqrt{2^{p+q}}}\mathbf{U}\mathbf{W}_q\mathbf{A_{vg}}\mathbf{S'_T}\mathbf{W}_p\mathbf{P}\mathbf{X}, \ \  m > n
		\end{cases}
	\end{equation}
	where $p$ is the minimum integer such that $2^p\ge m$, $q$ is the minimum integer such that $2^q\ge n$, $\mathbf{P}$ describes the zero-padding operation to make $\mathbf{X}$ multipliable by $\mathbf{W}_p$ or $\mathbf{W}_q$. $\mathbf{S'_T}$ represents the smooth-thresholding layer with DC channel excluded. $\mathbf{U}$ is unpadding function to make the dimension the same as $\mathbf{Z}$, and $\mathbf{Avg}$ is average pooling on the channel axis without the DC channel. Figure~\ref{fig: fwht} (a) shows an example to increase the number of channels and (b) another example to decrease the number of channels using FWHT layers, respectively.

	In consequence, if $d$ is the minimum integer such that $2^d$ is no less than the number of input channels, the trainable number of parameters in FWHT layers is no more than the $(2^d-1)$. The trainable parameters are only the threshold values in the smooth-thresholding operator. Hence, it is clear that FWHT layer requires significantly fewer parameters than the regular $1\times1$ convolution layer, which requires a different set of filter coefficients for each $1\times 1$ convolution.

\subsection{Block Walsh-Hadamard Transform layer}
    Although the FWHT layer is very efficient and can save a huge number of parameters, it requires zero padding before computing the FWHT when the size of the vector is not an integer power of 2. For example, if the input tensor has 384 channels, we have to pad 128 zeros to make it 512. These zeros contain no information but increase the number of trainable parameters significantly. Inspired by the block-division strategy of the Discrete Cosine Transform (DCT)-based JPEG image compression \cite{marcellin2000overview} method, we propose a layer in which we divide the input to small blocks and compute the Walsh-Hadamard transforms of small blocks of data. In this way, we do not need to pad a large number of zeros to the end of the input vector. For example, we divide the feature map into blocks of size 32 and compute 12 WHTs in an input tensor which has 384 channels. If necessary, we pad zeros to the end of the last block.
    
    The BWHT layer for channel expansion is described in Algorithm~\ref{al: BWHT layer expand}. When we want to increase the number of channels  by $t$ using the WHT of size $s$, we  overlap data blocks for a $c$-channel tensor. We change $c$ channels to $\frac{tc}{s}$ blocks and each block has $s$ channels. This is described in Algorithm~\ref{al: resample}. To achieve overlapping, we first create a $\lfloor \frac{tc}{s} \rfloor$-length arithmetic sequence $K=[k_0, k_1, ..., k_{\frac{tc}{s}-1}]$ from $0$ to $(c-s)$, then we use channels from index $\lfloor k_i \rfloor$ to index $(\lfloor k_i \rfloor+s-1)$ to build the $i$-th block. 
    After overlapping, we take the FWHT of each block among the last axis, perform smooth-thresholding with DC channel excluded, and compute the inverse FWHT.   Finally, we reshape the result to a tensor with $tc$ channels.
    
    \begin{algorithm}[htbp]
		\caption{The BWHT Layer for Channel Expansion}
		\begin{algorithmic}[1]
			\renewcommand{\algorithmicrequire}{\textbf{Input:}}
			\renewcommand{\algorithmicensure}{\textbf{Output:}}
			\REQUIRE Input tensor $\mathbf{X}\in\mathbb{R}^{n\times w\times h\times c}$, Hadamard size $s$
			\ENSURE  Output tensor $\mathbf{Z}\in\mathbb{R}^{n\times w\times h\times tc}$
			\STATE Resample $\mathbf{X}$ to get $\hat{\mathbf{X}}\in\mathbb{R}^{n\times w\times h\times \frac{tc}{s}\times{s}}$
			\STATE $\mathbf{Y} = \text{FWHT}(\hat{\mathbf{X}})\in\mathbb{R}^{n\times w\times h\times \frac{tc}{s}\times{s}}$
			\STATE $\hat{\mathbf{Y}} = \text{concat}(\mathbf{Y}[:, :, :, 0], \text{ST}(\mathbf{Y}[:, :, :, 1:]))\in\mathbb{R}^{n\times w\times h\times \frac{tc}{s}\times{s}}$
			\STATE $\hat{\mathbf{Z}} =\text{FWHT}(\hat{\mathbf{Y}})\in\mathbb{R}^{n\times w\times h\times \frac{tc}{s}\times{s}}$
			\STATE Reshape $\hat{\mathbf{Z}}$ to get $\mathbf{Z}\in\mathbb{R}^{n\times w\times h\times tc}$
			\RETURN $\mathbf{Z}$.\\
			Comments: FWHT($\cdot$) is the normalized fast Walsh-Hadamard transform on the last axis. Function concat($\cdot, \cdot$)  concatenates two tensors along the last axis. ST($\cdot$) performs smooth-thresholding. Index follows Python's rule.
		\end{algorithmic}
		\label{al: BWHT layer expand}
	\end{algorithm}
	
	\begin{algorithm}[htbp]
		\caption{Resampling}
		\begin{algorithmic}[1]
			\renewcommand{\algorithmicrequire}{\textbf{Input:}}
			\renewcommand{\algorithmicensure}{\textbf{Output:}}
			\REQUIRE Input tensor $\mathbf{X}\in\mathbb{R}^{n\times w\times h\times c}$, Hadamard size $s$
			\ENSURE  Output tensor $\hat{\mathbf{X}}\in\mathbb{R}^{n\times w\times h\times \frac{tc}{s}\times{s}}$
			\STATE $K=\lfloor\text{linspace}(0, c-s, \lfloor\frac{tc}{s}\rfloor)\rfloor$
			\FOR{$i$ in range($\frac{tc}{s}-1$)}
			    \STATE $\hat{\mathbf{X}}[:, :, :, i, :] = \mathbf{X}[:, :, :, K[i]:K[i]+s]$
			\ENDFOR 
			\RETURN $\hat{\mathbf{X}}$.\\
			Comments: $\lfloor\cdot\rfloor$ denotes the floor function. Function linspace($a, b, c$) creates a $c$-length arithmetic sequence from $a$ to $b$. Index follows Python's rule.
		\end{algorithmic}
		\label{al: resample}
	\end{algorithm}
	
	The BWHT layer for channel projection is described in Algorithm~\ref{al: BWHT layer project}.  We first divide the tensor into $\frac{tc}{s}$ blocks. Then, in each block, we perform an FWHT, smooth-thresholding, and compute the inverse FWHT. Finally, we reshape the output tensor to $tc$ channels and take an average pooling to $c$ channels. Unlike the FWHT for channel projection, we take the average pooling in the feature map domain instead of doing it in the WHT domain because the  transform size is much smaller in this case, and we want to keep as much information as possible.
	
	\begin{algorithm}[htbp]
		\caption{The BWHT Layer for Channel Projection}
		\begin{algorithmic}[1]
			\renewcommand{\algorithmicrequire}{\textbf{Input:}}
			\renewcommand{\algorithmicensure}{\textbf{Output:}}
			\REQUIRE Input tensor $\mathbf{X}\in\mathbb{R}^{n\times w\times h\times tc}$, Hadamard size $s$
			\ENSURE  Output tensor $\mathbf{Z}\in\mathbb{R}^{n\times w\times h\times c}$
			\STATE Reshape $\mathbf{X}$ to get $\hat{\mathbf{X}}\in\mathbb{R}^{n\times w\times h\times \frac{tc}{s}\times{s}}$
			\STATE $\mathbf{Y} = \text{FWHT}(\hat{\mathbf{X}})\in\mathbb{R}^{n\times w\times h\times \frac{tc}{s}\times{s}}$
			\STATE $\hat{\mathbf{Y}} = \text{concat}(\mathbf{Y}[:, :, :, 0], \text{ST}(\mathbf{Y}[:, :, :, 1:]))\in\mathbb{R}^{n\times w\times h\times \frac{tc}{s}\times{s}}$
			\STATE $\hat{\mathbf{Z}} =\text{FWHT}(\hat{\mathbf{Y}})\in\mathbb{R}^{n\times w\times h\times \frac{tc}{s}\times{s}}$
			\STATE Reshape $\hat{\mathbf{Z}}$ to get $\tilde{\mathbf{Z}}\in\mathbb{R}^{n\times w\times h\times tc}$
			\STATE $\mathbf{Z}=\text{avgpool}(\tilde{\mathbf{Z}}, t)\in\mathbb{R}^{n\times w\times h\times c}$
			\RETURN $\mathbf{Z}$.\\
			Comments: FWHT($\cdot$) is the normalized fast Walsh-Hadamard transform on the last axis. Function concat($\cdot, \cdot$) concatenates two tensors along the last axis. Function avgpool($\mathbf{A}, b$) is the average pooling on $\mathbf{A}$ with pooling size and strides are $b$. ST($\cdot$) performs smooth-thresholding. Index follows Python's rule.
		\end{algorithmic}
		\label{al: BWHT layer project}
	\end{algorithm}
    
\begin{figure}[htbp]
		\begin{center}
			\subfloat[BWHT layer for Channel Expansion]{\includegraphics[width=0.35\linewidth]{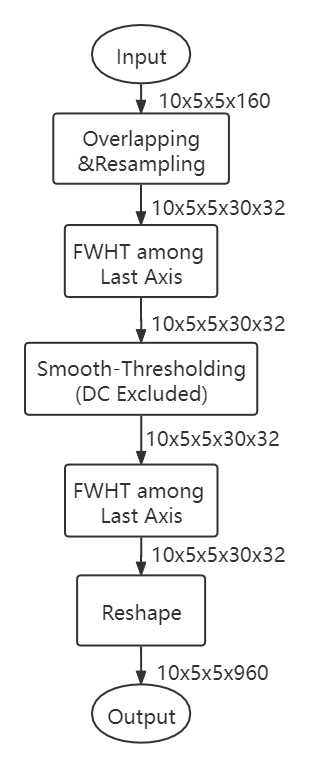}}
			\subfloat[BWHT layer for Channel Projection]{\includegraphics[width=0.35\linewidth]{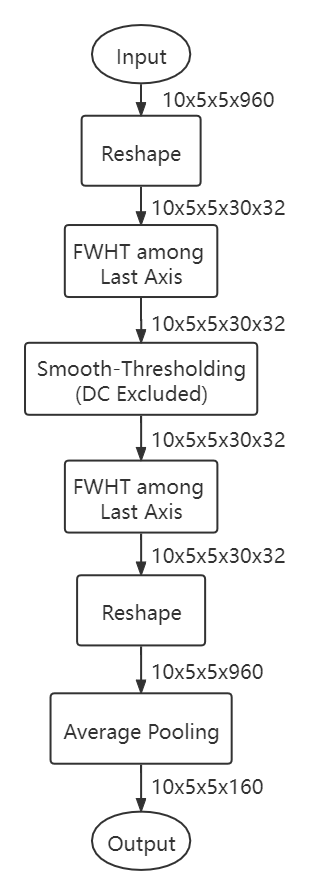}}
		\end{center}
		\caption{A BWHT layer (left) increases channel number of a tensor by 6 and a BWHT layer (right) decreases channel number of a tensor by $\frac{1}{6}$, respectively. There are only 31 trainable parameters in each BWHT layer because the WHT size is 32.}
		\label{fig: bwht}
	\end{figure}

    Figure~\ref{fig: bwht} shows an example to increase the number of channels and an example to decrease the number of channels using BWHT layers. The dimension change operation from $m$ dimensions to $n$ dimensions can be summarized as follows:
	\begin{equation}
		\mathbf{Z} = \begin{cases}
			\frac{1}{2^s}\mathbf{R}\mathbf{W}_q\mathbf{S'_T}\mathbf{W}_q\mathbf{G}\mathbf{X},  \ \ \ m \le n\\
			\frac{1}{\sqrt{2^s}}\mathbf{A_{vg}}\mathbf{R}\mathbf{W}_q\mathbf{S'_T}\mathbf{W}_p\mathbf{G}\mathbf{X}, \ \  m > n
		\end{cases}
	\end{equation}
    where $\mathbf{G}$ describes resampling or reshaping operations to divide $\mathbf{X}$ into blocks, $\mathbf{S'_T}$ is the smooth-thresholding layer with DC channel excluded, $\mathbf{R}$ represents the overlapping function to make the dimension the same as $\mathbf{Z}$, and $\mathbf{Avg}$ represents the average pooling on the channel axis.
    
	Consequently, compared to the FWHT layer with $2^d$ channels, we can reduce the trainable parameters from $(2^d-1)$ to $s$ by using the BWHT layer with a block size of $s$. For example, if there are 384 channels, a single FWHT layer contains 511 parameters while a BWHT layer only contains 31 parameters for a block size of $s=32$.
	
	We replace the 
	two $1\times 1$ convolution layers in each bottleneck layer of MobileNet-V2 with BWHT layers.
	
		
	
\subsection{2D Walsh-Hadamard Transform layer}
\begin{figure}[htbp]
		\begin{center}
			\subfloat[FNet in~\cite{lee2021fnet}\label{fig: FNet} ]{\includegraphics[width=0.4\linewidth]{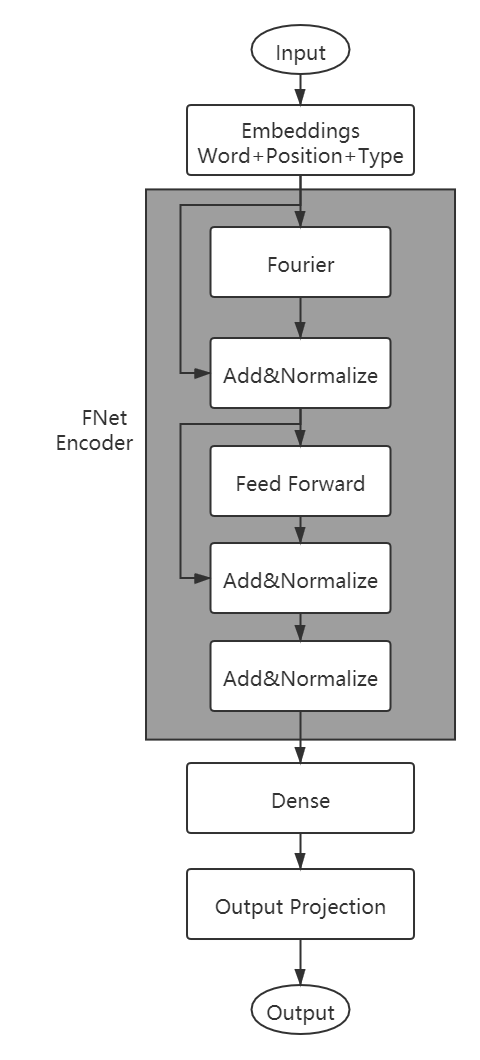}}
			\subfloat[2D-FWHT layer before the GAP layer\label{2D-FWHT before GAP}]{\includegraphics[width=0.4\linewidth]{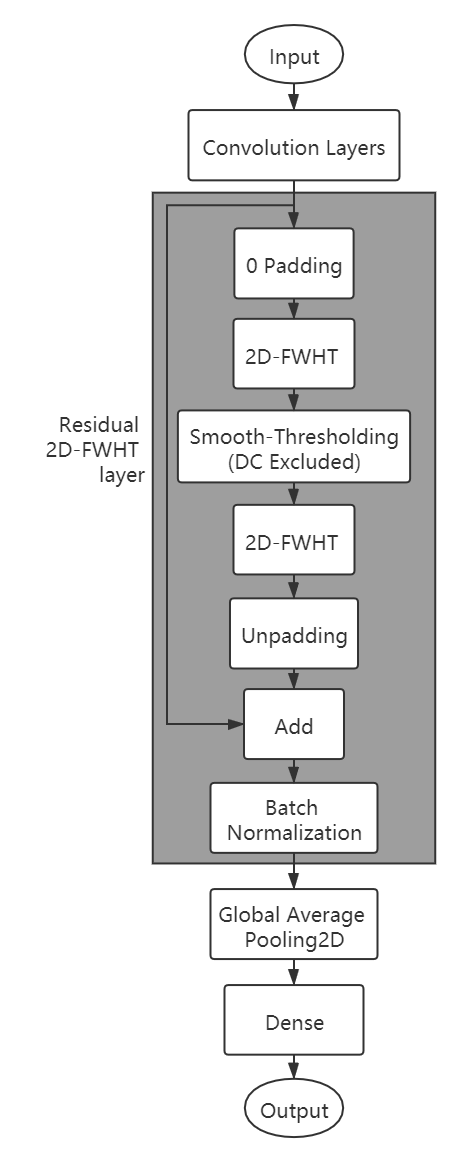}}
		\end{center}
		\caption{FNet and CNN with a 2D-FWHT layer before the GAP layer.}
		\label{fig: FNet and FWHT}
	\end{figure}

    In this section, we will introduce the two-dimensional (2D) Walsh-Hadamard Transform layers. In convolution neural networks, the global average pooling (GAP) and flatten layers are universally employed before the dense layers to reduce the dimension of the output tensor. 
    James Lee-Thorp \textit{et al.} proposed a Fourier Transform-based layer before the dense layer in their natural language processing network called FNet~\cite{lee2021fnet}.
    Inspired by their work, we design a 2D WHT layer that can be inserted before the GAP layer or the flatten layer as shown in Figure~\ref{2D-FWHT before GAP} and Equation~\ref{eq: 2D-FWHT}.
    \begin{equation}
        \mathbf{y}=\mathbf{W}\mathbf{S'_T}\mathbf{W}\mathbf{x}.
        \label{eq: 2D-FWHT}
    \end{equation}
    where, $\mathbf{W}$ denotes 2D-FWHT operation and $\mathbf{S'_T}$ represents the 2D smooth thresholding.
    
    In most convolution neural networks, the input tensor of the GAP and the flatten layer is in $\mathbb{R}^{N\times w\times h\times c}$, where $N$ is the batch size. Usually, $w=h<<c$. On the other hand, the output tensor of the GAP layer is in $\mathbb{R}^{N\times c}$ and the output tensor of the flatten layer is in $\mathbb{R}^{N\times whc}$. Therefore, the input of the GAP layer contains more information than the output of the GAP layer, and flatten layer reduces the 3D-spatial structure of the tensor. Due to these reasons, we insert the 2D-FWHT layer before the GAP layer or the flatten layer, and the 2D-FWHT layer aims to reinforce analysis among the width and height axes. 
    
\begin{figure}[htbp]
		\begin{center}
			\subfloat[Residual 2D-FWHT layer]{\includegraphics[width=0.35\linewidth]{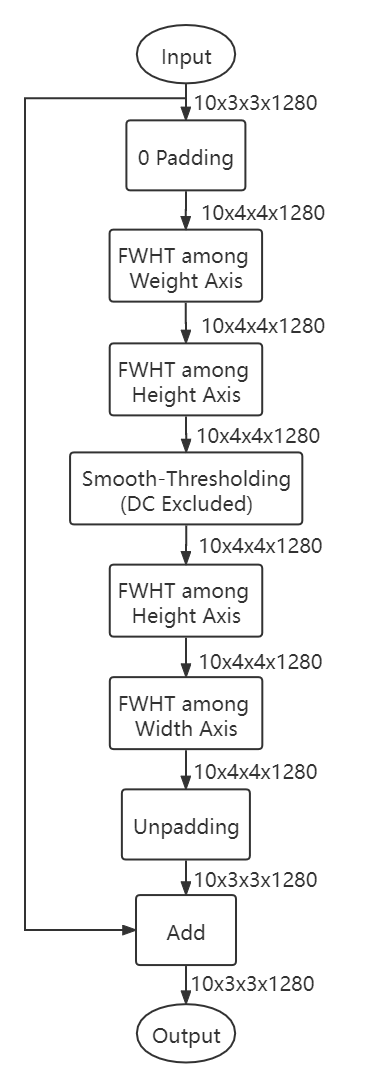}}
			\subfloat[Non-Residual 2D-FWHT layer]{\includegraphics[width=0.3\linewidth]{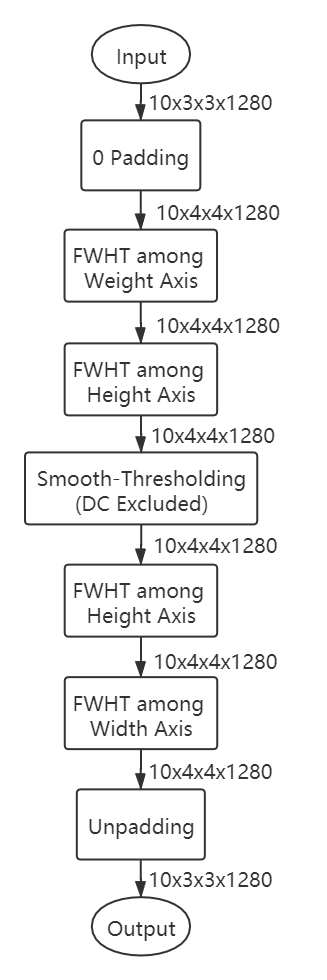}}
		\end{center}
		\caption{2D-FWHT layer examples. In each example, there are $4\times4=16$ trainable parameters while the one at DC is redundant. If weights are applied in the smooth-thresholding (details are in Section~\ref{sec: 2D Weighted Walsh-Hadamard Transform layer}, there are $4\times4=16$ extra trainable parameters. The residual one can be inserted before the GAP layer or the flatten layer to assist the dense layers, while the non-residual one can be employed to replace the 2D convolution layers and Squeeze-and-Excite layers.}
		\label{fig: 2dwht}
	\end{figure}

    As it is described in Algorithm~\ref{al: 2D-WHT layer} and Figure~\ref{fig: 2dwht}, we first pad zeros among the width and height axes to make the sizes of the transform powers of 2. Usually, $w$ and $h$ are very small integers if the convolution part is well-designed. Hence, the WHT size is not very large in this case (2, 4, or 8 is usually sufficient). Therefore, we do not apply block FWHT at this stage because we do not need to pad too many zeros. Then, we apply two 1D-FWHTs among the width and height axes separately to convert the tensor into the WH domain.  Next, we perform 2D-smooth-thresholding. There are $2^{p+q}$ trainable thresholding parameters, while the one at the DC channel is redundant but needed to perform Python's broadcasting. Each slice among the width and height axes will share a common threshold value. Afterwards, we reset the DC value to its original value and apply two 1D-FWHTs along the width and height axes separately to convert the tensor back to the feature domain. Finally, we un-pad the zeros from the result with the input tensor to get the output tensor. If it is with residual design, we will add the output with the input tensor to get the final output tensor.
    
    \begin{algorithm}[htbp]
		\caption{The 2D-FWHT Layer}
		\begin{algorithmic}[1]
			\renewcommand{\algorithmicrequire}{\textbf{Input:}}
			\renewcommand{\algorithmicensure}{\textbf{Output:}}
			\REQUIRE Input tensor $\mathbf{X}\in\mathbb{R}^{n\times w\times h\times c}$, Hadamard size $s$
			\ENSURE  Output tensor $\mathbf{Z}\in\mathbb{R}^{n\times w\times h\times c}$
			\STATE find minimum $p, q\in \mathbb{N}$, s.t. $2^p\ge w, 2^q \ge h$
			\STATE $\hat{\mathbf{X}}=\text{pad2}(\mathbf{X}, 2^p-w, 2^q-h)\in\mathbb{R}^{n\times 2^p\times 2^q\times c}$
			\STATE $\mathbf{Y} =\text{FWHT2}(\hat{\mathbf{X}})\in\mathbb{R}^{n\times 2^p\times 2^q\times c}$
			\STATE $\hat{\mathbf{Y}} =\text{ST2}(\mathbf{Y})\in\mathbb{R}^{n\times 2^p\times 2^q\times c}$
			\STATE$\hat{\mathbf{Y}}[:, 0, 0, :] = \mathbf{Y}[:, 0, 0, :]$
			\STATE $\hat{\mathbf{Z}} =\text{FWHT2}(\hat{\mathbf{Y}})\in\mathbb{R}^{n\times 2^p\times 2^q\times c}$
			\IF{Residual is applied}
			    \STATE $\mathbf{Z} = \hat{\mathbf{Z}}[:, :w, :h, :] + \mathbf{X}$
			\ELSE
			    \STATE $\mathbf{Z} = \hat{\mathbf{Z}}[:, :w, :h, :]$
			\ENDIF
			\\
			Comments: Function pad2($\mathbf{A}, b, c$) pads $b$ zeros among the width axis and $c$ zeros among the height axis of tensor $\mathbf{A}$. FWHT2($\cdot$) is the normalized 2D-fast Walsh-Hadamard transform on the width and height axes (applying two 1D-FWHTs separately for efficiency). ST2($\cdot$) performs 2D-smooth-thresholding. Index follows Python's rule.
		\end{algorithmic}
		\label{al: 2D-WHT layer}
	\end{algorithm}

	We also use the 2D-FWHT layers  to replace the regular 2D-convolution layers. In this case,  it does not change the number of channels due to the fact that it takes no transformation among the channel axis. Therefore, if we want to change the number of channels, we can apply a 1D-BWHT first then apply a 1D-BWHT layer to change the number of channels, then apply a 2D-FWHT layer. If the weight and height of the input tensor are not very small, we can also apply the block division as the BWHT layer to avoid padding too many zeros. 
	
	\subsection{2D Weighted Walsh-Hadamard Transform layer}\label{sec: 2D Weighted Walsh-Hadamard Transform layer}
	Convolution in "time" domain can be implemented using multiplicative weights in the Fourier domain. However, Fourier transform requires complex arithmetic. In this paper, we replaced the Fourier Trannsform with the binary WHT. 
	The weighted 2D WHT layer uses multiplicative weights in the transform domain. In the weighted-WHT layer we perform both regular "filtering" and nonlinear "denoising" in the transform domain.
	
	To improve the accuracy, we use the weighted smooth-thresholding function, in which a trainable weight is applied:
    \begin{equation}
		y = \text{S}_{WT}' (x) = \tanh(x)(|vx|-T)_+
		\label{eq: Weighted Smooth-Thresholding}
	\end{equation}
	with constraint $v\ge0$, where $v$ is a trainable multiplicative weight representing a WHT domain multiplication. 
	The corresponding partial derivatives are
	\begin{equation}
		\frac{\partial(\tanh(x)(|vx|-T)_+)}{\partial T} = \begin{cases}
			\tanh(x)|x|, & |vx| > T\\
			0, & |vx| \le T\\
		\end{cases}
		\label{eq: dSWTdW}
	\end{equation}
	and 
	\begin{equation}
		\frac{\partial(\tanh(x)(|vx|-T)_+)}{\partial T} = \begin{cases}
			-\tanh(x), & |vx| > T\\
			0, & |vx| \le T\\
		\end{cases}
		\label{eq: dSWTdT}
	\end{equation}
    When $|vx| > T$, Eq. (\ref{eq: dSWTdW}) holds because when $v\ge0$, $\tanh(x)x\text{sign}(vx)=\tanh(x)x\text{sign}(x)=	\tanh(x)|x|$. We don't multiply $v$ and $x$ in $\tanh(\cdot)$ in Eq.~(\ref{eq: Weighted Smooth-Thresholding}) to avoid quadratic term of $v$ in the derivatives. In this way, Eq.~(\ref{eq: 2D-FWHT}) can be rewritten as:
    
    \begin{equation}
        \mathbf{y}=\mathbf{W}\mathbf{S'_{WT}}\mathbf{W}\mathbf{x}.
        \label{eq: 2D-WFWHT}
    \end{equation}
    where, $\mathbf{W}$ denotes 2D-FWHT and $\mathbf{S'_{WT}}$ denotes 2D weighted smooth thresholding. 
    We initialize all weights $v=1$ in the 2D weighted smooth-thresholding layer. We initialize the threshold value $T=0$ at the DC channel. The threshold values at other AC channels are initialized as positive values. In weighted-WHT layer, the number of trainable parameters are doubled but the number of parameters in the 2D-FWHT layer is still significantly smaller than a regular convolution layer.
    For example, if the input and the output are in $\mathbb{R}^{3\times3\times1280}$ and the WHT size is 4, there are only $4\times4+4\times4=32$ trainable parameters in weighted 2D-FWHT layer, but $3\times3$ 2D convolution layer requires $3\times3\times1280=11520$ trainable parameters. 
    

\section{Experimental Results}

Our training and testing experiments are carried on an HP-Z820 workstation with 2 Intel Xeon E5-2695 v2 CPUs, 2 NVIDIA RTX A4000 GPUs, and 128GB RAM. Speed tests are also carried out using a NVIDIA Jetson Nano board.  The code is written in TensorFlow-Keras in Python 3.

First, we will compare the accuracy of MobileNet-V2 based neural networks on the Fashion MNIST dataset 
and CIFAR-10 dataset 
. Then, we will further investigate 2D-FWHT in MobileNet-V3 and ResNet. Finally, we will perform a speed test on our 2D-FWHT layer and the regular $3\times3$ convolution layer. 

\subsection{BWHT in MobileNet-V2}
In MobileNet-V2 experiments, we use ImageNet-pretrained MobileNet-V2 model with 1.0 depth~\cite{Mobilenet_web}. As it is shown in Table~\ref{tab: model}, to build the fine-tuned MobileNet-V2, we replace layers after the Global Average Pooling (GAP) layer with a dropout layer and a dense layer as TensorFlow official transfer learning demo \cite{Transfer_learning_web}. Since the minimum input of MobileNet-V2 ImageNet-pretrained models is $96\times96\times3$, we resize the images to this resolution in our experiments. Because images in Fashion MNIST dataset are in gray-scale, after interpolating, we copy the values two times to convert them to $96\times96\times3$ format. 
	
	\begin{table}[htbp]
	\caption{Structure of fine-tuned MobileNet-V2 (baseline). $t$, $c$, $n$ and $s$ represent expansion factor, channel, repeat time and stride \cite{sandler2018mobilenetv2}. Initial weights before dropout are from ImageNet checkpoint float\_v2\_1.0\_96 in \cite{Mobilenet_web}.}
		\begin{center}
			\begin{tabular}{cccccc}
    \toprule
				Input & Operator & $t$ & $c$ & $n$ & $s$\\
    \midrule
				$96^2\times 3$&Conv2D&-&32&1&2\\
				$48^2\times 32$&Bottleneck&1&16&1&1\\
				$48^2\times 16$&Bottleneck&6&24&2&2\\
				$24^2\times 24$&Bottleneck&6&32&3&2\\
				$12^2\times 32$&Bottleneck&6&64&4&2\\
				$6^2\times 64$&Bottleneck&6&96&3&1\\
				$6^2\times 96$&Bottleneck&6&160&3&2\\
				$3^2\times 160$&Bottleneck&6&320&1&1\\
				$3^2\times 320$&Conv2D&-&1280&1&1\\
				$3^2\times 1280$&AvgPool&-&-&1&-\\
				$1280$&Dropout (rate=0.2)&-&-&1&-\\
				$1280$&Dense (units=10)&-&-&1&-\\
    \bottomrule
			\end{tabular}
		\end{center}	
		\label{tab: model}
	\end{table}


Results of MobileNet-V2 on Fashion MNIST and CIFAR-10 are shown in Tables~\ref{tab: Fashion MNIST} and ~\ref{tab: CIFAR-10} respectively. The number of parameters is counted by TensorFlow API ``model.summary()", and the non-trainable parameters are from the batch normalization layers. We first investigate the 2D-FWHT layer before the GAP layer. We name the models as ``2D-FWHT before GAP". We add one residual 2D-FWHT layer followed by a batch normalization layer. They bring an extra 2576 more trainable parameters (only 16 from the 2D-FWHT layer) and only 2560 non-trainable parameters but improve the accuracy by 0.27\% and 0.47\% on Fashion MNIST and CIFAR-10, respectively. The number of trainable parameters increases only by 0.12\%. 

\begin{table}[htbp]
  \caption{MobileNet-V2 Fashion MNIST Result}
  \label{tab: Fashion MNIST}
  \begin{tabular}{ccccc}
    \toprule
    Model&Trainable&Non-Trainable&Trainable Parameters &Accuracy\\
    &Parameters&Parameters&Reduction Ratio&\\
    \midrule
    Fine-tuned MobileNet-V2 (baseline)&2,236,682&34,112&-&95.37\%\\
    2D-FWHT before GAP&2,239,258&36,672&-&95.64\%\\
    $\frac{1}{2}$ 1x1 conv. changed (FWHT) \cite{pan2021fast}&288,983&34,112&87.08\%&94.50\%\\
    $\frac{1}{2}$ 1x1 conv. changed (BWHT) &273,177&34,112&87.79\%&94.45\%\\
    \textbf{$\frac{1}{2}$ 1x1 conv. changed (BWHT)}&\multirow{2}{*}{\textbf{275,753}}&\multirow{2}{*}{\textbf{36,372}}&\multirow{2}{*}{\textbf{87.67\%}}&\multirow{2}{*}{\textbf{94.75\%}}\\
    \textbf{+2D-FWHT before GAP}&&&&\\
    \bottomrule
\end{tabular}
\end{table}

\begin{table}[htbp]
  \caption{MobileNet-V2 CIFAR-10 Result}
  \label{tab: CIFAR-10}
  \begin{tabular}{ccccc}
    \toprule
    Model&Trainable&Non-Trainable&Trainable Parameters &Accuracy\\
    &Parameters&Parameters&Reduction Ratio&\\
    \midrule
    Baseline MobileNet-V2 model$\mathbf{^a}$ in \cite{ayi2020rmnv2}&2.2378M&-&-&94.3\%\\
	RMNv2$\mathbf{^b}$ \cite{ayi2020rmnv2}&1.0691M&-&52.22\%&92.4\%\\
    \midrule
    Fine-tuned MobileNet-V2 (baseline)&2,236,682&34,112&-&95.21\%\\
    Fourier Layer~\cite{lee2021fnet} before GAP&2,239,242&36,672&-&95.38\%\\
    2D-FWHT before GAP&2,239,258&36,672&-&95.68\%\\
    $\frac{1}{3}$ 1x1 conv. changed (FWHT) \cite{pan2021fast}&506,094&34,112&77.37\%&93.14\%\\
    $\frac{1}{3}$ 1x1 conv. changed (BWHT)&494,175&34,112&77.91\%&93.22\%\\
    \textbf{$\frac{1}{3}$ 1x1 conv. changed (BWHT)}&\multirow{2}{*}{\textbf{496,751}}&\multirow{2}{*}{\textbf{36,672}}&\multirow{2}{*}{\textbf{77.79\%}}&\multirow{2}{*}{\textbf{93.46\%}}\\
    \textbf{+2D-FWHT before GAP}&&&&\\
    \bottomrule
    \multicolumn{5}{l}{Parameters and accuracy of baseline model$\mathbf{^a}$ and RMNv2$\mathbf{^b}$ are from Table 3 in \cite{ayi2020rmnv2}.}
\end{tabular}
\end{table}

\begin{figure}[htbp]
		\begin{center}
			\subfloat[MobileNet-V2 bottleneck \cite{sandler2018mobilenetv2}]{\includegraphics[width=0.3\linewidth]{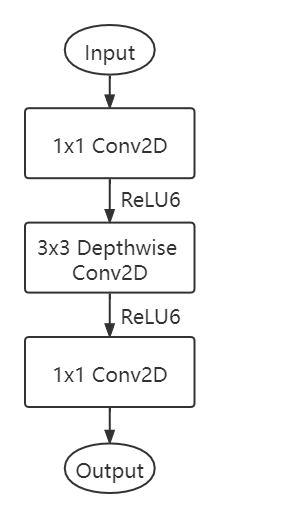}}
			\subfloat[Our version with 1D-BWHT layer]{\includegraphics[width=0.3\linewidth]{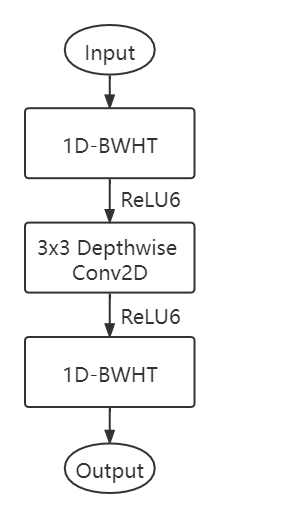}}
		\end{center}
			\caption{MobileNet-V2 bottleneck and our version with 1D-BWHT layer.}
		\label{fig: mobilenet_bottleneck}
	\end{figure}

We then beat the FWHT results in \cite{pan2021fast} which do not apply block division. As Figure~\ref{fig: mobilenet_bottleneck} shows, we first revise the $1\times1$ convolution layers. The models are named as ``1x1 conv. changed". 
The fraction at the beginning of the model's name denotes how many bottleneck layers are changed. For instance, $\frac{1}{2}$ means we change the last half bottleneck layers. Since the final convolution layer is also $1\times1$, we replace it with a BWHT layer for channel expansion. 

 We also compared our results with Reduced Mobilenet-V2 (RMNv2)~\cite{ayi2020rmnv2} in CIFAR-10 experiment. They slim MobileNet-V2 by replacing bottleneck layers with a novel block called HetConv blocks. Their trimming method is still based on the convolution. When we change $\frac{1}{3}$ convolution by BWHT layer and add an extra 2D-FWHT layer before the GAP layer, we reduce more parameters than (77.79\% VS 52.22\%) ~\cite{ayi2020rmnv2} with a smaller accuracy loss (1.75\% VS 1.9\%).

 Moreover, we compare our results with the FNet Fourier layer~\cite{lee2021fnet} in CIFAR-10 experiment. The FNet Fourier layer can be computed as 
 \begin{equation}
     \mathbf{Y} = \mathbf{X}+\mathfrak{R}(\mathcal{F}_w(\mathcal{F}_h(\mathbf{X})))
 \end{equation}
 where the input tensor is $\mathbf{X}$ and the output tensor is $\mathbf{Y}$. $\mathcal{F}_w(\cdot)$ and $\mathcal{F}_h(\cdot)$ are the Fast Fourier Transform (FFT) among the weight and the height. $\mathfrak{R}(\cdot)$ is the real part of the tensor. We try adding one FNet layer before the GAP layer to compare the result of adding one 2D-FWHT layer before the GAP layer, and our accuracy is 0.30\% higher than the accuracy of the FNet. This is because the FNet can only keep the real part of the tensor. In another words the information in the imaginary part is simply discarded. On the other hand, the Walsh-Hadamard Transform is a binary transform. Therefore, there is no information loss due to the Walsh-Hadamard Transform. 

According to the above experiments, we have the following observations:
\begin{itemize}
    \item In general, the weighted 2D-FWHT before the GAP layer improves the accuracy of an image recognition network. For example, it improves the accuracy of the fine-tuned MoblieNet-V2 by 0.27\% in Fashion MNIST and 0.47\% in CIFAR-10. Furthermore, it increases the accuracy of the model ``$\frac{1}{2}$ 1x1 conv. changed BWHT" by 0.4\% on Fashion MNIST and accuracy of model ``$\frac{1}{3}$ 1x1 conv. changed BWHT" on CIFAR-10 by 0.24\%, respectively.
    \item The 1D-BWHT layer provides better results than the single block 1D-FWHT layer in terms of both the accuracy and the number of parameters when the number of channels is not the power of 2. Although on Fashion MNIST, the 1D-FWHT model reaches higher accuracy, the BWHT models reach higher accuracy on CIFAR-10. CIFAR-10 is a more difficult dataset so it is more convincing than the Fashion MNIST. Plus, in each case, the BWHT models have fewer parameters than the 1D FWHT models because the BWHT layer avoids padding 0s and it uses a smaller WHT size.

\end{itemize}

\subsection{2D-FWHT in MobileNet-V3}\label{sec: 2D-FWHT in MobileNet-V3}
\begin{figure}[htbp]
		\begin{center}
			\subfloat[MobileNet-V3 bottleneck]{\includegraphics[width=0.4\linewidth]{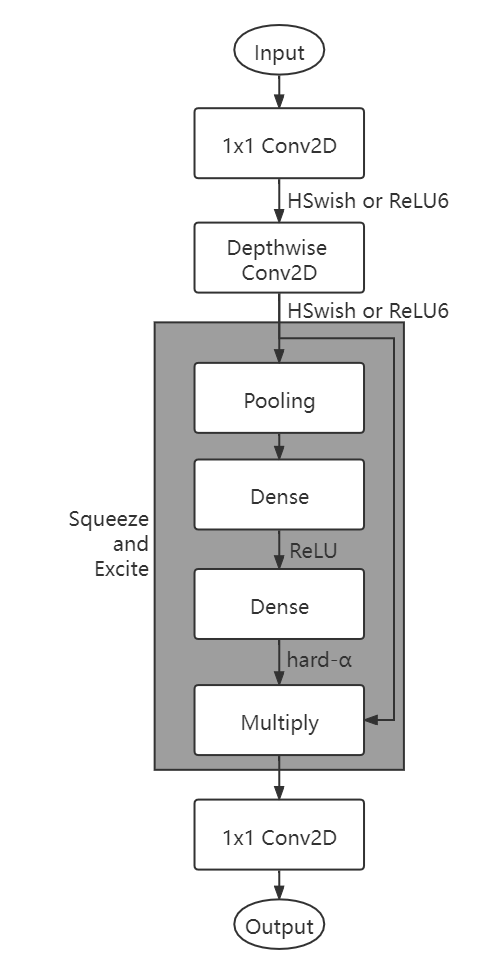}}
			\subfloat[Our version with 2D-FWHT layer]{\includegraphics[width=0.3\linewidth]{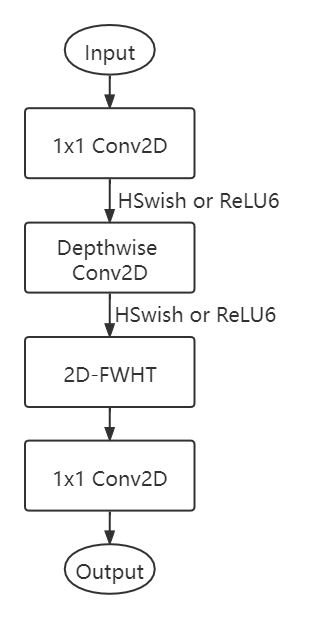}}
		\end{center}
		\caption{MobileNet-V3 bottleneck~\cite{howard2019searching} and our version with 2D-FWHT layer.}
		\label{fig: mobilenetv3_bottleneck}
	\end{figure}

In this section, we use MobileNet-V3-Large \cite{howard2019searching} on CIFAR-100 dataset and Tiny ImageNet to show the advantage of the 2D-FWHT layer. As Figure~\ref{fig: mobilenetv3_bottleneck} shows, compared to the MobileNet-V2 bottleneck block, the MobileNet-V3 bottleneck block contains one Squeeze-and-Excite layer after the depthwise convolution layer. In this section, we replace each Squeeze-and-Excite layer in the final $\frac{1}{3}$ bottleneck blocks with the 2D-FWHT layer. As Table~\ref{tab: model_v3} shows, the base model is MobileNet-V3-large with the weights in the early layers are pretrained on the ImageNet dataset. Due to the input size of public ImageNet-pretrained MobileNet-V3 models are all $224\times224$ but CIFAR image size is 32x32, we upsample the images to $224\times224$.

\begin{table}[htbp]
	\caption{Structure of the fine-tuned MobileNet-V3-Large. The table follows Table 1 in \cite{howard2019searching}. Initial weights before dropout are from ImageNet checkpoint ``Large dm=1 (float)" in \cite{Mobilenet_web}. SE denotes
whether there is a Squeeze-And-Excite in that block. ``NL" denotes
the type of nonlinearity used. Here, ``HS" denotes h-swish and ``RE"
denotes ReLU. ``s" denotes
stride.}
		\begin{center}
			\begin{tabular}{ccccccc}
    \toprule
	Input & Operator & exp size & \#out & SE & NL&s\\
    \midrule
	$224^2\times3$&Conv2d&-&16&-&HS&2\\
	$112^2\times16$&bneck, 3x3&16&16&N&RE&1\\
	$112^2\times16$&bneck, 3x3&64&24&N&RE&2\\
	$56^2\times24$&bneck, 3x3&72&24&N&RE&1\\
	$56^2\times24$&bneck, 5x5&72&40&Y&RE&2\\
	$28^2\times40$&bneck, 5x5&120&40&Y&RE&1\\
	$28^2\times40$&bneck, 5x5&120&40&Y&RE&1\\
	$28^2\times40$&bneck, 3x3&240&80&N&HS&2\\
	$14^2\times80$&bneck, 3x3&200&80&N&HS&1\\
	$14^2\times80$&bneck, 3x3&184&80&N&HS&1\\
	$14^2\times80$&bneck, 3x3&184&80&N&HS&1\\
	$14^2\times80$&bneck, 3x3&480&112&Y&HS&1\\
	$14^2\times112$&bneck, 3x3&672&112&Y&HS&1\\
	$14^2\times112$&bneck, 5x5&672&160&Y&HS&2\\
	$7^2\times160$&bneck, 5x5&960&160&Y&HS&1\\
	$7^2\times160$&bneck, 5x5&960&160&Y&HS&1\\
	$7^2\times160$&conv2d, 1x1&-&960&-&HS&1\\
	$7^2\times960$&AvgPool&-&-&-&-&1\\
	960&Dropout(rate=0.2)&-&-&-&-&-\\
	960&Dense(units=10 for CIFAR-10, 100 for CIFAR-100, 200 for Tiny ImageNet)&-&-&-&-&-\\
    \bottomrule
			\end{tabular}
		\end{center}	
		\label{tab: model_v3}
	\end{table}

As Table~\ref{tab: MobileNetV3_CIFAR100} shows, we reduce 48.55\% trainable parameters with only a 0.33\% accuracy loss by changing one-third of ($\frac{1}{3}$) Squeeze-and-Excite layers and applying one 2D-FWHT layer before the GAP layer. Compared to the cases without applying weights in the 2D smooth-thresholding, all cases with weights get improved accuracy. Specifically, when we insert a 2D-FWHT layer in the 2D-smooth-threshold before the GAP layer, the network with weights in the 2D smooth-thresholding reaches 0.51\% accuracy higher than the network without weights in the 2D smooth-thresholding. When we change $\frac{1}{3}$ Squeeze-and-Excite layers, the network with weights in the 2D smooth-thresholding reaches 0.15\% accuracy higher than the network without weights in the 2D smooth-thresholding. In each pair, the parameters amount is almost the same respectively. Therefore, weighted smooth-thresholding is superior to non-weighted smooth-thresholding in accuracy.

We also  compare the WHT layer with FNet Fourier layer~\cite{lee2021fnet}. When we insert a  layer before the GAP layer, the model with our 2D-FWHT reaches a 0.62\% accuracy higher than the model with FNet Fourier layer. In addition, when we change $\frac{1}{3}$ Squeeze-and-Excite layers, the model with 2D-FWHT reaches a 0.29\% accuracy higher than the model with FNet Fourier layer.

\begin{table}[htbp]
\centering
  \caption{MobileNet-V3-Large CIFAR-100 Result}
  \label{tab: MobileNetV3_CIFAR100}
  \begin{tabular}{cccccc}
        \toprule
    Model&Weights in 2D&Trainable&Non-Trainable&Trainable Parameters&Accuracy\\
    &Thresholding&Parameters&Parameters&Reduction Ratio&\\
        \midrule
    Fine-tuned model (Baseline)&-&3,068,052&24,400&-&78.25\%\\
    Fourier layer~\cite{lee2021fnet} before GAP&-&3,069,972&26,320&-&80.01\%\\
    2D-FWHT before GAP&N&3,070,036&26,320&-&80.22\%\\
    2D-FWHT before GAP&Y&3,070,100&26,320&-&80.73\%\\
    $\frac{1}{3}$ S\&E changed (Fourier~\cite{lee2021fnet})&-&1,574,988&24,400&&77.20\%\\
    $\frac{1}{3}$ S\&E changed (2D-FWHT)&N&1,575,692&24,400&48.64\%&77.34\%\\
    $\frac{1}{3}$ S\&E changed (2D-FWHT)&Y &1,576,396&24,400&48.62\%&77.49\%\\
    \textbf{$\frac{1}{3}$ S\&E changed+2D-}&\multirow{2}{*}{\textbf{Y}}&\multirow{ 2}{*}{\textbf{1,578,444}}&\multirow{2}{*}{\textbf{26,320}}&\multirow{2}{*}{\textbf{48.55\%}}&\multirow{2}{*}{\textbf{77.92\%}}\\
    \textbf{FWHT before GAP}&&&&&\\
    \bottomrule
    \multicolumn{6}{l}{S\&E is the Squeeze-and-Excite layer.}
\end{tabular}
\end{table} 

Furthermore, we also try our best model on CIFAR-10 and Tiny ImageNet. We replace each Squeeze-and-Excite layer in the final $\frac{1}{3}$ bottleneck blocks by the 2D-FWHT layer and insert one 2D-FWHT layer before the GAP layer. As Table~\ref{tab: MobileNetV3_CIFAR10} shows, our model has 49.96\% less trainable parameters than the fine-tuned MobileNet-V3-Large model and the accuracy is only 1.20\% lower on the CIFAR-10. As Table~\ref{tab: MobileNetV3_TinyImageNet} shows, our model has 47.08\% less trainable parameters than the fine-tuned MobileNet-V3-Large model and the accuracy is only 1.59\% lower on the Tiny ImageNet.

\begin{table}[htbp]
\centering
  \caption{MobileNet-V3-Large CIFAR-10 Result}
  \label{tab: MobileNetV3_CIFAR10}
  \begin{tabular}{cccccc}
        \toprule
    Model&Weights in 2D&Trainable&Non-Trainable&Trainable Parameters&Accuracy\\
    &Thresholding&Parameters&Parameters&Reduction Ratio&\\
        \midrule
    Fine-tuned model (Baseline)&-&2,981,562&24,400&-&95.52\%\\
    \textbf{$\frac{1}{3}$ S\&E changed+2D-}&\multirow{2}{*}{\textbf{Y}}&\multirow{ 2}{*}{\textbf{1,491,954}}&\multirow{2}{*}{\textbf{26,320}}&\multirow{2}{*}{\textbf{49.96\%}}&\multirow{2}{*}{\textbf{94.32\%}}\\
    \textbf{FWHT before GAP}&&&&&\\
    \bottomrule
    \multicolumn{6}{l}{S\&E is the Squeeze-and-Excite layer.}
\end{tabular}
\end{table} 

\begin{table}[htbp]
\centering
  \caption{MobileNet-V3-Large Tiny ImageNet Result}
  \label{tab: MobileNetV3_TinyImageNet}
  \begin{tabular}{cccccc}
        \toprule
    Model&Weights in 2D&Trainable&Non-Trainable&Trainable Parameters&Accuracy\\
    &Thresholding&Parameters&Parameters&Reduction Ratio&\\
        \midrule
    Fine-tuned model (Baseline)&-&3,164,152&24,400&-&64.37\%\\
    \textbf{$\frac{1}{3}$ S\&E changed+2D-}&\multirow{2}{*}{\textbf{Y}}&\multirow{ 2}{*}{\textbf{1,674,544}}&\multirow{2}{*}{\textbf{26,320}}&\multirow{2}{*}{\textbf{47.08\%}}&\multirow{2}{*}{\textbf{62.78\%}}\\
    \textbf{FWHT before GAP}&&&&&\\
    \bottomrule
    \multicolumn{6}{l}{S\&E is the Squeeze-and-Excite layer.}
\end{tabular}
\end{table} 

\subsection{2D-FWHT in ResNet}\label{sec: 2D-FWHT in ResNet}
\begin{figure}[htbp]
		\begin{center}
			\subfloat[Residual block \cite{he2016deep}\label{fig: residual}]{\includegraphics[width=0.3\linewidth]{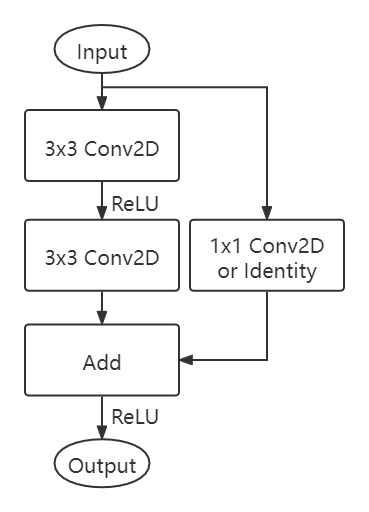}}
			\subfloat[Completely revised residual block\label{fig: residual_completely_revised}]{\includegraphics[width=0.3\linewidth]{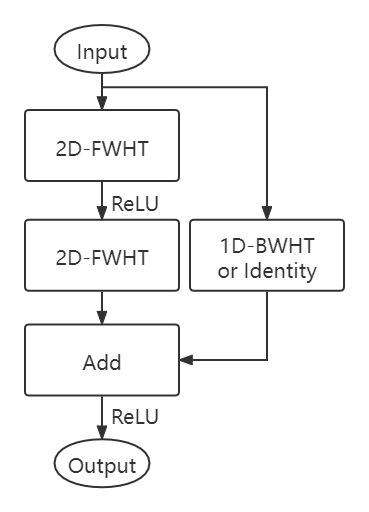}}
			\subfloat[Partially revised residual block\label{fig: residual_partially_revised}]{\includegraphics[width=0.3\linewidth]{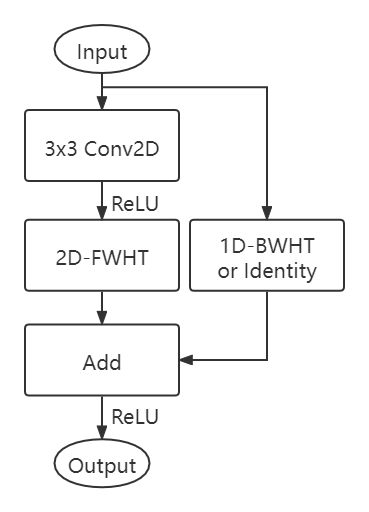}}
		\end{center}
		\caption{Residual Block and our revised versions. $1\times1$ convolution or 1D-BWHT layer is applied if the number of channels increases.}
		\label{fig: residual block}
	\end{figure}

In this section, we will investigate 2D-FWHT in ResNet. ResNet is built with standard convolution layers \cite{he2016deep}. It does not use any depthwise convolution layer. Therefore, we can employ the ResNet to show the advantage of our 2D-FWHT layer. We do not use any pre-trained weights but we initialize the weights with Kaiming He's initialization \cite{he2015delving}. We first build ResNet-20 as shown in Table~\ref{tab: resnet-20} and Figure~\ref{fig: residual}. As Table~\ref{tab: ResNet20_CIFAR10} shows, it reaches an accuracy of 91.60\% on CIFAR-10 with 273,066 trainable parameters. Then, we revise all residual blocks by replacing $3\times3$ convolution layers with 2D-FWHT layers and replacing $1\times1$ convolution layers with 1D-BWHT layers as shown in Figure~\ref{fig: residual_completely_revised}. Here we apply 2D-FWHT layers without residual design because the blocks already contain the residual design. Since the dimension numbers are already in the power of 2, we do not need to pad any zeros before computing the Walsh-Hadamard transforms. We do not implement an additional 2D-FWHT layer before the GAP layer because the input of the GAP layer is already from a 2D-FWHT layer. In this way, we save 95.76\% parameters. In other words, there are virtually no parameters left before the GAP layers. Note that the dense layer contains 650 parameters, the batch normalization layers have 2,752 parameters, and the first convolution layer holds 448 parameters. There are only 7,740 parameters in the 2D-FWHT layers and the 1D-BWHT layers. With an extremely low number of trainable parameters, the network still reaches 60.47\% accuracy. We then revise the residual partially as shown in Figure~\ref{fig: residual_partially_revised}. We retain the first $3\times3$ convolution layer in each residual block and change other convolution layers. In this way, we save 52.76\% trainable parameters with only a 1.72\% accuracy loss. Finally, we add weights in the 2D smooth-threshold, and we reduce 51.26\% trainable parameters with only a 1.48\% accuracy loss.

\begin{table}[htbp]
	\caption{Structure of ResNet-20 for CIFAR-10. Building blocks are shown in brackets, with the numbers of blocks stacked. Downsampling is performed by conv3\_1 and conv4\_1 with a stride of 2. Batch normalization layer is applied after each convolution layer.}
    \begin{center}
    \begin{tabular}{ccc}
        \toprule
		Layer Name&Output Shape&Implementation Details\\
        \midrule
		Input&$32\times32\times3$&-\\
		Conv1&$32\times32\times16$&$3\times3, 16$\\
		Conv2\_x&$32\times32\times16$&$\left[ \begin{array}{c} 3\times3, 16  \\ 3\times3, 16 \end{array}\right]\times 3$\\
		Conv3\_x&$16\times16\times32$&$\left[ \begin{array}{c} 3\times3, 32  \\ 3\times3, 32 \end{array}\right]\times 3$\\
		Conv4\_x&$8\times8\times64$&$\left[ \begin{array}{c} 3\times3, 32  \\ 3\times3, 64 \end{array}\right]\times 3$\\
		GAP&$64$&Global Average Pooling2D\\
		Output&$10$&Dense(unit = 10)\\
    \bottomrule
	\end{tabular}
\end{center}	
\label{tab: resnet-20}
	\end{table}
	
\begin{table}[htbp]
  \caption{ResNet-20 CIFAR-10 Result}
  \label{tab: ResNet20_CIFAR10}
  \begin{tabular}{cccccc}
    \toprule
    Model&Weights in 2D&Trainable&Non-Trainable&Trainable Param &Accuracy\\
    &Thresholding&Parameters&Parameters&Reduction Ratio&\\
    \midrule
    ResNet-20 in \cite{he2016deep}&-&0.27M&-&-&91.25\%\\
    ResNet-20 in our trial (baseline)&-&273,066&1,376&-&91.60\%\\
    2D-FWHT before GAP&N&273,258&1,504&-&91.74\%\\
    2D-FWHT before GAP&Y&273,322&1,504&-&91.75\%\\
    Residual blocks completely revised&N&11,590&1,376&95.76\%&60.47\%\\
    Residual blocks partially revised&N&129,000&1,376&52.76\%&89.88\%\\
    \textbf{Residual blocks partially revised }&\textbf{Y}&\textbf{133,082}&\textbf{1,376}&\textbf{51.26\%}&\textbf{90.12\%}\\
    \bottomrule
\end{tabular}
\end{table}

We further explore ResNet-34 on the Tiny ImageNet. The base model is defined as Table~\ref{tab: resnet-34}. As it is shown in Table~\ref{tab: ResNet34_TinyImageNet}, we reduce 53.66\% trainable parameters with only 0.89\% accuracy loss without weights in 2D smooth-thresholding and 53.37\% trainable parameters with only 0.72\% accuracy loss with weights in 2D smooth-thresholding.

\begin{table}[htbp]
	\caption{Structure of ResNet-34 for Tiny ImageNet. Building blocks are shown in brackets, with the numbers of blocks stacked. Downsampling is performed by conv3\_1, conv4\_1 and conv5\_1 with a stride of 2. Batch normalization layer is applied after each convolution layer.}
    \begin{center}
    \begin{tabular}{ccc}
        \toprule
		Layer Name&Output Shape&Implementation Details\\
        \midrule
		Input&$64\times64\times3$&-\\
		Conv1&$64\times64\times64$&$3\times3, 64$\\
		Conv2\_x&$64\times64\times64$&$\left[ \begin{array}{c} 3\times3, 64  \\ 3\times3, 64 \end{array}\right]\times 3$\\
		Conv3\_x&$32\times32\times128$&$\left[ \begin{array}{c} 3\times3, 128  \\ 3\times3, 128 \end{array}\right]\times 4$\\
		Conv4\_x&$16\times16\times256$&$\left[ \begin{array}{c} 3\times3, 256  \\ 3\times3, 256 \end{array}\right]\times 6$\\
		Conv5\_x&$8\times8\times512$&$\left[ \begin{array}{c} 3\times3, 512  \\ 3\times3, 512 \end{array}\right]\times 3$\\		
		GAP&$512$&Global Average Pooling2D\\
		Output&$200$&Dense(unit = 200)\\
    \bottomrule
	\end{tabular}
\end{center}	
\label{tab: resnet-34}
	\end{table}

\begin{table}[htbp]
  \caption{ResNet-34 Tiny ImageNet Result}
  \label{tab: ResNet34_TinyImageNet}
  \begin{tabular}{cccccc}
    \toprule
    Model&Weights in 2D&Trainable&Non-Trainable&Trainable Param &Accuracy\\
    &Thresholding&Parameters&Parameters&Reduction Ratio&\\
    \midrule
    ResNet-34 (baseline)&-&21,386,312&15,232&-&53.06\%\\
    Residual blocks partially revised&N&9,910,565&15,232&53.66\%&52.17\%\\
    \textbf{Residual blocks partially revised }&\textbf{Y}&\textbf{9,928,677}&\textbf{15,232}&\textbf{53.57\%}&\textbf{52.34\%}\\
    \bottomrule
\end{tabular}
\end{table}

\subsection{Speed and Memory Tests}
In our early work \cite{pan2021fast}, we have shown that the 1D-FWHT layer runs about 2 times as fast as the $1\times1$ convolution layer on the NVIDIA Jetson Nano (4GB version) when the input and the output are both in $\mathbb{R}^{10\times32\times32\times1024}$ and the input tensor is initialized randomly. Now we will compare the 2D-FWHT layer versus the $3\times3$ convolution layer to show the speed advantage of our 2D-FWHT layer for the ResNet and other deep CNNs with regular convolution layers.

In this experiment, we set the input and the output are both in $\mathbb{R}^{10\times8\times8\times1024}$. We use the same laptop (with Intel Core i7-7700HQ CPU, NVIDIA GTX-1060 GPU with Max-Q design, and 16GB DDR4 RAM) as \cite{pan2021fast} to run the TensorFlow PB model, and we deploy the TFLite model to the same NVIDIA Jetson Nano. As it is stated in Table~\ref{tab: speed}, the 2D-FWHT layer runs about 24 times as fast as the $3\times3$ convolution layer on the NVIDIA Jetson Nano. There are two reasons why the time difference on NVIDIA Jetson Nano is much more significant than on the laptop. One reason is that the TFLite model on the NVIDIA Jetson Nano is optimized by TensorFlow for ARM embedded devices, while the PB model on the laptop is not optimized as much as the TFLite model. The other reason is that the Windows backend apps slow down the system, while the NVIDIA Jetson Nano system is ``clean". 

We also compare the 2D-FWHT layer with the Squeeze-and-Excite layer. The squeeze ratio is $\frac{1}{4}$ as the Squeeze-and-Excite layer in the MobileNet-V3.  We achieve comparable times in squeeze-and-excite layers. This is because current processors cannot distinguish between multiplication by $\pm 1$ and a real number.

\begin{table}[htbp]
	\caption{Speed test}
		\begin{center}
			\begin{tabular}{cccc}
                \toprule
				Device&$3\times3$ Conv2D&\textbf{2D-FWHT}&Squeeze-and-Excite (Ratio = $\frac{1}{4}$)\\
                \midrule
                Laptop (GPU)&0.0508 S&\textbf{0.0459 S}&0.0449 S\\
				Laptop (CPU)&0.0768 S&\textbf{0.0479 S}&0.0424 S\\
				NVIDIA Jetson Nano&1.8861 S&\textbf{0.0776 S}&0.0519 S\\
                \bottomrule
                \multicolumn{4}{l}{Input tensor and output tensor are in $\mathbb{R}^{10\times8\times8\times1024}$. code is available at~\cite{Hadamard_speed_code}.}
			\end{tabular}
		\end{center}	
		\label{tab: speed}
	\end{table}

\begin{figure}[htbp]
    \centering
    \includegraphics[width=0.6\linewidth]{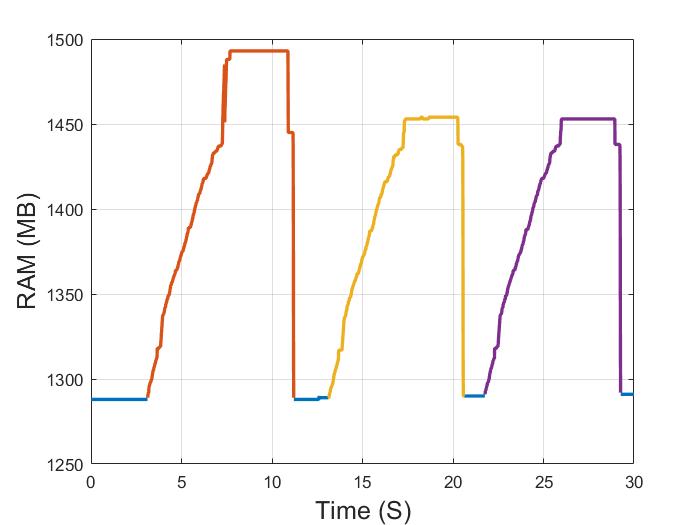}
    \caption{Memory test on an NVIDIA Jetson Nano. From left to right, three peaks denote the inference memory of the $3\times3$ Conv2D (Max: 1493 MB), 2D-FWHT (Max: 1454 MB), and Squeeze-and-Excite layers  (Max: 1453 MB). In each peak, after the layer is loaded, we run the inference code in a while-loop until we reach a steady state. RAM usage is 1288 MB when the device is vacant. Input and output tensors are in $\mathbb{R}^{10\times8\times8\times1024}$.}
    \label{fig: memory}
\end{figure}

In addition, We record the RAM usage of the three layers on the NVIDIA Jetson Nano. According to Figure~\ref{fig: memory}, the 2D-FWHT layer and the Squeeze-and-Excite layer have close RAM usages, and they require about 40MB (19.51\%) less memory than the $3\times3$ Conv2D in the inference. 


\section{Conclusions}
In this paper, we proposed 1D and 2D Walsh-Hadamard Transform (WHT)-based binary layers to replace $1\times1$ and $3 \times 3$ convolution layers in deep neural networks. The 2D-FWHT layer can also be used to replace the Squeeze-and-Excite layers common to many publicly available networks including the MobileNet-V3 \cite{howard2019searching}. 

We implement WHT in blocks of data using its $O(m\log_2m)$ Fast-WHT (FWHT) algorithm, where $m$ is the number of elements in each block.
Compared to the FWHT layer in ~\cite{pan2021fast} which does not apply block by block computation, the Block-WHT (BWHT) layer avoids zero paddings, if the transform size $m$ is chosen wisely as a power of 2, and our new approach can save more trainable parameters compared to our earlier paper~\cite{pan2021fast}. 

In addition, we propose a residual 2D-FWHT layer that can be easily inserted before the global average pooling (GAP) layer (or the flatten layer) to assist the dense layers. By applying the 1D-BWHT layers and a 2D-FWHT layer before the GAP layer on the MobileNet-V2, we save 87.67\% trainable parameters with only a 0.62\% accuracy loss on the Fashion MNIST and 77.79\% trainable parameters with a 1.75\% accuracy decrease on the CIFAR-10. Similarly, by inserting one 2D-FWHT layer before the GAP layer and replacing Squeeze-and-Excite layers with the 2D-FWHT layers on the MobileNet-V3-Large, we save 49.96\% trainable parameters with a 1.20\% accuracy loss on the CIFAR-10, 48.55\% trainable parameters with a 0.33\% accuracy loss on the CIFAR-100, and 47.08\% trainable parameters with a 1.59\% accuracy decrease on the Tiny ImageNet.

By inserting a 2D-FWHT layer before the GAP layer in MobileNet-V3-Large, we achieve an increase of 2.48\% in accuracy on CIFAR-100 with a slight increase (0.07\%) in trainable parameters. 

Furthermore, the non-residual 2D-FWHT layer can be also used as a replacement for the $3\times3$ convolution layers and the Squeeze-and-Excite layers. We can reduce nearly half trainable parameters for MobileNet-V3-Large with a negligible accuracy loss (0.33\% accuracy loss on the CIFAR-100 dataset).


As a final point, we compare the speed of our 2D-FWHT layer with the regular $3\times3$ 2D convolution layer. In an NVIDIA Jetson Nano board, our 2D-FWHT layer runs about 24 times as fast as the regular 2D convolution layer with 19.51\% less RAM usage.

\bibliographystyle{unsrt}
\bibliography{sample-base}

\end{document}